\documentclass{article}
\pdfoutput=1

\usepackage{microtype}
\usepackage{graphicx}
\usepackage{subfig}
\usepackage{booktabs} 

\usepackage{amsmath,amsfonts,bm}
\usepackage{amsthm}
\usepackage{thmtools,thm-restate}

\def\eqref#1{Equation~\ref{#1}}

\def\1{\bm{1}}

\DeclareMathAlphabet{\mathsfit}{\encodingdefault}{\sfdefault}{m}{sl}
\SetMathAlphabet{\mathsfit}{bold}{\encodingdefault}{\sfdefault}{bx}{n}

\usepackage[title]{appendix}

\usepackage{amsfonts}  
\usepackage{amsmath}

\usepackage{hyperref}

\usepackage{adjustbox}
\usepackage[textwidth=1.3in]{todonotes}
\usepackage{tikz}
\usepackage{pgfplots}
\usepackage{multirow}
\usepackage{titletoc}
\usepackage{minitoc}
\usepackage{multirow}
 \usepackage{wrapfig,booktabs}
\usepackage{algorithm}
\usepackage{algorithmic} 
\usepackage{xspace} 
\usepackage{subfig}
\usepackage{array}
\newcolumntype{H}{>{\setbox0=\hbox\bgroup}c<{\egroup}@{}}
\newcommand{\TheMethod}{GIFT\xspace}

\usepackage[preprint]{neurips_2021}

\usepackage[utf8]{inputenc} 
\usepackage[T1]{fontenc}    
\usepackage{hyperref} 
\usepackage{url}            
\usepackage{booktabs}       
\usepackage{amsfonts}       
\usepackage{nicefrac}       
\usepackage{microtype}      
\usepackage{xcolor}         

\usepackage{selectp}
\title{Gradual Domain Adaptation in the Wild: \\ When Intermediate Distributions are Absent}

\author{%
 Samira Abnar, Rianne van den Berg, Golnaz Ghiasi, Mostafa Dehghani, \\ \textbf{Nal Kalchbrenner, Hanie Sedghi} \\
 Google Research, Brain team \\
 \texttt{\{samiraabnar, riannevdberg, hsedghi\}@google.com} \\
}

\begin{document}

\maketitle

\begin{abstract}
We focus on the problem of domain adaptation when the goal is shifting the model towards the target distribution, rather than learning domain invariant representations.
It has been shown that under the following two assumptions: (a) access to samples from intermediate distributions, and (b) samples being annotated with the amount of change from the source distribution, self-training can be successfully applied on gradually shifted samples to adapt the model toward the target distribution.  We hypothesize having (a) is enough to enable iterative self-training to slowly adapt the model to the target distribution, by making use of an implicit curriculum. In the case where (a) does not hold, we observe that iterative self-training falls short. We propose GIFT, a method that creates virtual samples from intermediate distributions by interpolating representations of examples from source and target domains.
We evaluate an iterative-self-training method on datasets with natural distribution shifts, and show that when applied on top of other domain adaptation methods, it improves the performance of the model on the target dataset. We run an analysis on a synthetic dataset to show that in the presence of (a) iterative-self-training naturally forms a curriculum of samples. Furthermore, we show that when (a) does not hold, GIFT performs better than iterative self-training. 

\end{abstract}

\section{Introduction}

Machine learning algorithms are notorious for not being robust to changes in the environment and their performance often drops significantly when there is a big shift in the data distribution~\citep{taori2020measuring, hendrycks2020many, koh2020wilds}.
For learning algorithms to be robust to the changes in the distribution, they either need to learn representations that are invariant to the shift or they should update their parameters to be more aligned with the new distribution. While unsupervised domain adaptation techniques commonly rely on learning domain invariant representations, our focus is on shifting the model towards the target distribution and we consider the unsupervised setting where we do not have access to labels on the target domain. 

We look at this problem through the lens of curriculum learning. Curriculum learning~\citep{elman1993learning,sanger1994neural, bengio2009curriculum} suggests presenting easier samples early on in the training process and gradually increasing the difficulty. In unsupervised domain adaptation, this is equivalent to gradually changing the distribution from the source domain to the target domain, i.e., getting the model to adapt to intermediate distributions before being exposed to the target domain.  
\citet{kumar2020understanding} show that if learning algorithms are exposed to gradual changes in the data distribution under a self-training regime,  the generalization gap (from source to target distribution) will be much lower.  However, the problem remains unresolved when we do not have access to the intermediate steps of the distribution shift, either because they do not exist or because of limitations in the data collection process. 

\begin{figure}
    \centering
    \includegraphics[width=0.5\textwidth]{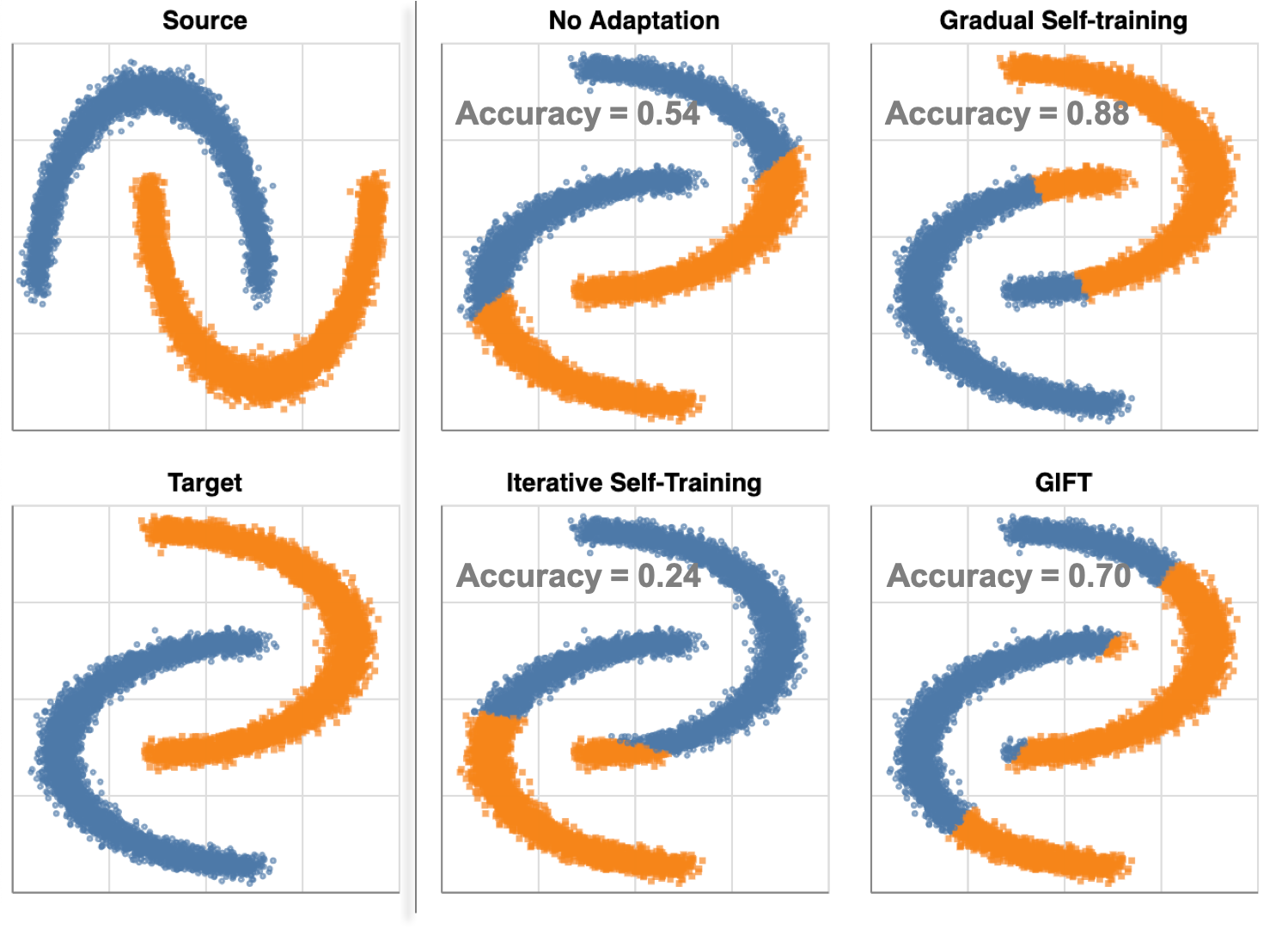}
    \caption{Demonstrating the power of self-training for shifting the model on the Two-moon dataset. Here the target data is the ($90^{\circ}$) rotated  version of the source data. In this example, it is not possible to achieve a good performance on both source and target at the same time, since they have conflicting labels for similar inputs. While the performance of the model trained only on source is around $\sim50\%$ on the target data, if we have access to ground truth intermediate steps we can improve this number by $\sim40\%$. In the absence of samples from intermediate distributions,  GIFT can increase the performance by $\sim20\%$. For this example we use an MLP with one hidden layer and Relu activation function. The code for replicating this example is provided at \url{https://github.com/samiraabnar/Gift/blob/main/notebooks/noisy_two_moon.ipynb}.
    }
    \label{fig:two_moon_example}
\end{figure}

Findings from \citet{kumar2020understanding} suggest that to be able to adapt the model by self-training, two conditions should be satisfied:
(a) Access to samples from intermediate distributions between source and target;
(b) Access to information about the amount of shift for each sample.
We argue that while (b) is hard to achieve in practice, (a) may be the case in many real-world scenarios.
We hypothesize that if (a) holds, i.e., the target distribution supports intermediate steps to some degree,
iterative self-training, i.e., applying self-training iteratively while filtering examples based on the confidence of the model,  can incorporate an implicit curriculum based on the confidence of the model for each example from the target domain, and this curriculum helps the model to gradually adapt to the target domain. 
Furthermore, for cases where (a) does not hold, we propose GIFT (Gradual Interpolation of Features toward Target). GIFT creates virtual examples from intermediate distributions by linearly interpolating between source and target data in the input and feature space of a neural network.
To gradually increase the difficulty of the virtual samples, the linear interpolation coefficient changes such that the samples start at the source distribution and gradually move towards the target distribution during training. Figure~\ref{fig:two_moon_example}, demonstrates how GIFT can improve iterative-self-training in an example with a two-moon dataset.

We evaluate iterative self-training for unsupervised domain adaptation on datasets with natural distribution shifts and show that combined with other domain adaptation methods, it improves performance on the target dataset. On a synthetic benchmark, we show that in the absence of (a), GIFT performs better than iterative self-training.
By tracking the accuracy and confidence of a model on different subsets of the target distribution we show that both \TheMethod and iterative self-training are indeed doing curriculum learning. \TheMethod has two advantages over iterative self-training: (1) It works better when the number of training iterations is limited, (2) it works better when the target distribution is not diverse enough to include a mixture of easy and hard examples.

\section{Self-training for unsupervised domain adaptation}

Being able to properly handle distribution shift is one of the primary concerns of machine learning algorithms. A common setting where this problem is considered is unsupervised domain adaptation, where we have access to labeled data from one or multiple source domains, and unlabeled data from the target domain. Let $(X^s, Y^s)$ denote the labeled data in the source domain, where $X^s \in \mathbb{R}^{n_s \times d}$ is the sample input
matrix and $Y^s \in \mathbb{R}^{n_s \times k}$ is the corresponding label matrix. Let 
$X^t \in \mathbb{R}^{n_t \times d} $ denote the unlabeled target domain data. We assume some underlying or common features exist between source and target, while there is substantial distribution shift between the two domains. The goal is to bridge the
domain difference and learn a good classifier for the target domain. The first assumption is needed for domain adaptation to be successful~\citep{ben2008does}. The latter emphasizes that a model trained on the source domain presents a noticeable performance gap on the target domain and hence needs to be adapted to the target distribution.

Self-training uses a teacher model that is trained on source domain $P_s$, to produce pseudo labels together with confidence scores on unlabeled data from the target domain $P_t$ and uses these predictions to train a student model. 
In this paper we examine how the confidence scores of the teacher model guide the student model to adapt to the target domain, and whether we can devise supplementary schemes that help the student adapt better.

\section{\TheMethod: Self-training with Gradual Interpolations}

The curriculum learning paradigm suggests presenting the easier samples early in the training and gradually increasing the difficulty of the samples~\cite{elman1993learning,sanger1994neural, bengio2009curriculum}. In unsupervised domain adaptation, this is equivalent to gradually changing the distribution from the source domain to the target domain, i.e., getting the model to adapt to intermediate distributions before getting exposed to the target domain. 

If in addition to labeled data from the source domain and unlabeled data from the target domain, we also have access to the intermediate distributions, i.e., unlabeled examples from data distributions between source and target, we can use them to boost the performance of the model on the target domain, by applying self-training in a gradual manner. 
More precisely, we can apply self-training in an iterative manner and ensure a small shift between the distribution that the model is trained on in the previous iteration and the self-training examples at current iteration.

This approach results in a self-training procedure with a more effective adaptation mechanism.
\citet{kumar2020understanding} show that this gradual adaptation leads to a lower error bound on the target domain.
In order for this approach to be applicable, we need samples from intermediate steps that are annotated with the amount of shift from the source distribution. 
However, in practice, we either do not have access to samples of intermediate distributions or such examples do not exist. Even if they do, it is less likely that they are annotated with the degree of shift.

In order to circumvent the issue of lack of samples from intermediate distributions, we create the virtual samples from intermediate distributions by interpolating the input and hidden representations of the data from the source domain $P_s$ and target domain $P_t$. Namely, let $M_\phi$ be a neural network model trained on source data and let $z_i^s$ correspond to the representation of input $x_i^s \in X^s$ from the source domain. We choose $x_j^t$ as a sample from the target domain (we explain the procedure to pick $x_j^t \in X^t$ below). Let $z_j^s$ correspond to the representation of input $x_j^t \in X^t$ and $\lambda \in [0,1]$. We generate 
\begin{equation}
  \label{eqn:interpolate}   \widehat{z}_{ij} = (1 - \lambda) z_i^s + \lambda z_j^t,\qquad  \lambda \in [0,1]
\end{equation}
as a sample representation of a virtual intermediate distribution. Note that we do not seek to find a corresponding input and we only focus on the feature space. Gradually increasing $\lambda$ from 0 to 1 corresponds to feature representations that change from corresponding source distribution representations ($\lambda = 0$) to the target distribution representation ($\lambda=1$ ).
\begin{algorithm}[tb]
	\caption{\TheMethod: Gradual Interpolation}
	 $P_s$, : source dataset;
	 $P_t$: target dataset\\
   	 $\mathcal{M}$: Neural net (maps input to predictions). \\
   	 $\mathcal{M}^{:L}$: partial neural net (maps input to features of layer L)\\
  	 $\mathcal{M}^{L:}$:  partial neural net (maps features of layer L to predictions)\\
   	  $\phi$: student neural net parameters;
   	  $\theta$: teacher neural net parameters\\
	 $\delta$: step size for interpolation coefficient $\lambda$.\\
	 $N$: number of training iterations per teacher update. \\
	 $\mathrm{\alpha}$: confidence threshold (percentage of examples) \\
	 Notation: $[m]= \{1,...m\}$.\\
	\label{alg:main}
	\begin{algorithmic}[1]
    \STATE $\mathcal{M}_{\phi} \gets \mathcal{M}_{\theta}$, $\lambda \gets 0$
	\WHILE {$\lambda <= 1$ }
		\FOR {$step \in N$}
		\STATE $(x^s_{[m]}, y^s_{[m]}) \sim P_s$ 
		\STATE $x^t_{[m]} \sim P_t$
		\STATE $z^s_{[m]} \gets \mathcal{M}^{:L}_\phi(x^s_{[m]})$
		\STATE $z^t_{[m]} \gets \mathcal{M}^{:L}_\theta(x^t_{[m]})$
		\STATE $y^t_{[m]} \gets \mathcal{M}_\theta(x^t_{[m]})$
		\STATE $\mathrm{index}_s, \mathrm{index}_t \! \leftarrow \! \textbf{align}(y^s_{[m]}, y^t_{[m]})$ (Algorithm~\ref{alg:align})
		\STATE $\widehat{z}_{[m]} \gets (1 -\lambda) \times z^s_{\mathrm{index}_s} + \lambda \times z^t_{\mathrm{index}_t} $
		\STATE $\widehat{y}_{[m]}  \gets \mathcal{M}^{L:}_\theta(\widehat{z}_{[m]})$
		\STATE $\mathrm{conf\_ranks} \gets rank(max(\widehat{y}_{[m]} ) - min(\widehat{y}_{\{...\}}))$
		\STATE $\mathrm{conf\_indices} \gets \mathrm{conf\_ranks}[:\mathrm{\alpha}]$
        \STATE $\widehat{z}_{[m]} , \widehat{y}_{[m]}  \gets \widehat{z}[\mathrm{conf\_indices}],  \widehat{y}[\mathrm{conf\_indices}]$
		\STATE  Update $\mathcal{M}_\phi$ to fit  $(\widehat{z}_{[m]} ,  \widehat{y}_{[m]} )$
		\ENDFOR
			\STATE $\lambda \gets \textbf{lambda\_scheduler}(\lambda, \delta)$
			\STATE $\mathcal{M}_\theta \gets \mathcal{M}_\phi$
	\ENDWHILE
	\end{algorithmic} 
\end{algorithm} 

\begin{algorithm}[tb]
\caption{Align: Label-based Random Alignment}
\label{alg:align}
\begin{algorithmic}[1]
\STATE {\bfseries Input:} {$y^s_{[m]}, \hat{y}^t_{[m]}$}
\STATE {\bfseries Results:} {$\mathrm{index}_s, \mathrm{index}_t$}
\STATE $\mathrm{index_s} \gets [1, 2, ..., len(y^s_{[m]})]$
\STATE $\mathrm{index_t} \gets []$
\FOR{$i \in \mathrm{index_s}$}
    \STATE $\mathrm{indices} \gets [1, 2, ..., len(\hat{y}^t_{[m]})]$
    \STATE $\mathrm{shuffled\_indices} \gets permute(\mathrm{indices})$
    \STATE $\mathrm{index} \gets argmax(y^s[i] == \hat{y}^t[\mathrm{shuffled\_indices}])$ 
    \STATE $\mathrm{index}_t.append(\mathrm{shuffled\_indices}[\mathrm{index}])$
\ENDFOR
\end{algorithmic}	
\end{algorithm}

To choose which example from the source is interpolated with which example from the target, we either randomly align the samples, or apply a cost-based alignment method based on the $L_2$ distance of representations, and the similarity/equality of predicted labels. To apply a cost-based alignment method, we use the Sinkhorn matching algorithm~\citep{sinkhorn_1966, MAL-073} to approximate the alignment with the lowest cost. 
Although iterative cost-based alignment methods such as Sinkhorn come at a computational cost as compared to random alignment, we have observed that random alignment without taking (pseudo) labels into account leads to worse performance.
To find a better trade-off between alignment cost and performance, in our experiments, we also tried a non-iterative heuristic (pseudo) label-based random alignment, where we randomly align examples that have the same (pseudo) labels. This is shown in Algorithm~\ref{alg:align}. In our experiments, we report the result from cost-based alignment.  We observed that pseudo-random alignment and cost-based alignment lead to more or less similar results and in practice we could simply use the pseudo random alignments.
It is important to note that both for cost-based and pseudo-random alignment we use the pseudo labels predicted by the teacher model, since we do not have access to ground truth labels for data points from the target domain.

In standard iterative self-training~\citep{Habrard2013IterativeSD}, the data used in each self-training iteration is a subset of the target distribution, whereas in \TheMethod we apply the iterative self-training procedure on virtual intermediate distribution representations to gradually adapt the model to the target domain. We start by using a teacher model trained on the source distribution and move to train the student model in the representation space. The self-training procedure at each step proceeds by assigning pseudo labels to virtual intermediate representations that are generated by~\eqref{eqn:interpolate}. Next, the student model is updated using pseudo-labeled representations. 
Then the student becomes the teacher for the next iteration and the procedure continues. We start with $\lambda = 0$, which generates representations from the source domain. At each iteration of self-training, we increase the value of $\lambda$ to generate virtual representations that are closer to the target distribution and hence gradually move the model toward the target distribution ($\lambda = 1$). The details are shown in Algorithm~\ref{alg:main}.
Similar to standard iterative self-training, we assign a confidence score to each pseudo-labeled data point and only update the student model with high-confidence pseudo-labeled data. As a confidence score, we use the gap between the highest and lowest logit for each sample, as proposed in~\cite{kumar2020understanding}.

\section{Experiments}
In our experiments, first, we examine the power of iterative self-training and GIFT for unsupervised domain adaptation on datasets with natural distribution shift. 
Then, we use a dataset with synthetic shift, where we can track the amount of shift and control the gap between the source and targets distribution. We demonstrate the implicit curriculum followed by the iterative self-training method and show that when condition (a) does not hold, GIFT performs better than iterative self-training.

The training setup that we employ has three main phases:
(i) Pre-training: Pre-training the model on a large scale dataset (ImageNet-1k in our experiments)
(ii) Fine-tuning: Training the pre-trained model on a labeled source domain with or without leveraging samples from an unlabeled target domain. 
(iii) Adaptation: Applying a self-training based approach to shift the model towards the unlabeled target domain.
In the fine-tuning stage, step (ii), we compare 
(A) simple fine-tuning on the source domain with standard augmentation techniques, i.e., random flip and random crop, 
(B) fine-tuning on the source domain with variants of Mixup~\citep{zhang2018mixup} as an augmentation technique, and (c) Domain Adversarial training that uses unlabeled examples from the target domain to learn domain invariant representations~\citep{ganin2016domain}.
In step (iii), the adaptation phase, we compare three self-training based strategies: (1) one-step self-training,  (2) iterative self-training and (3) GIFT,
which is similar to iterative self-training except that during the intermediate iterations the model is trained with samples from virtual intermediate states instead of actual samples from the target distribution.

\subsection{Benchmarks with Natural Distribution Shift}

\begin{table}[]
    \centering
    \caption{Accuracy on target domain on benchmarks with natural distribution shift. For the experiments in this table we use a ResNet-101 pretrained on ImageNet-1k.}
    \adjustbox{width=\textwidth}{
    \begin{tabular}{l c H c c c}
    \toprule
         \multirow{2}{*}{\textbf{Method}} & \textbf{Office31} & \textbf{DomainNet} & \textbf{FMoW} & \textbf{Camelyon17}  & \textbf{Imagenet-R}  \\ 
         & amazon $\rightarrow$ webcam & real $\rightarrow$ painting & train $\rightarrow$ test\_ood & 0 $\rightarrow$ 3 & Imagenet $\rightarrow$ Imagenet-R\\
         \midrule
         Fine-tuned on Source (A) & 0.696 & domainnet & 0.502  & 0.807 & 0.385\\
         Mixup-Convex (B) & 0.703 & domainnet & \textbf{0.525} & 0.591 & - \\
         Mixup-Wasserstein (C) & 0.727 & domainnet & 0.499 & 0.896 & - \\
         DANN (D) & \textbf{0.750} & 0.542 & 0.505 & \textbf{0.934} & - \\ \hline
         Best (A, B, C, D) + Self-training & \textbf{0.772} & & 0.530 & 0.962 & 0.417\\
         Best (A, B, C, D) + Iterative Self-training & 0.771 &  & \textbf{0.539} & 0.966 & 0.448\\
         Best (A, B, C, D) + GIFT & 0.761 & & \textbf{0.539} & \textbf{0.973} & \textbf{0.462} \\\bottomrule
    \end{tabular}}
    \label{tab:results_natural_shift}
\end{table}

We report results on four different datasets with natural distribution shifts: FMoW~\citep{christie2018functional} and Camelyon17~\citep{bandi2018detection} from the WILDS benchmark~\citep{koh2020wilds}, and Office 31~\citep{office31}, which is widely used as a domain adaptation benchmark, as well as Imagenet-R~\citep{hendrycks2020many}.
Here we briefly introduce each of these datasets:

\textbf{FMoW} is a variant of the Functional Map of the World dataset that contains satellite imagery of earth. The images belong to 62 building or land use categories, and the domain represents both the year the image was taken as well as its geographical region. Here we only address the domain shift problem over time. For the adaptation phase we use unlabeled samples from the out-of-distribution test split of the dataset. 

\textbf{Camelyon17} is the patch-based variant of Camelyon17~\citep{bandi2018detection} with hospitals (${0, 1, 2, 3, 4}$) as domains. The task is to predict if a given region of tissue contains a tumor. We use $(0,1,2)$ as source and $3$ as target. 


\textbf{Office31} contains images of objects from 31  categories in three domains: Amazon, DSLR and Webcam. We use Amazon and Webcam as the source and target domain respectively.

\textbf{ImageNet-R}enditions is introduced as a benchmark to measure generalization to different abstract visual renditions. The images in this dataset are a subset of line drawings from \cite{wang2019learning} and images retrieved from Flicker. It has 30,000 examples for 200  ImageNet classes. Approaches such as \texttt{DeepAugment+AugMix} and \texttt{DeepAugment} achieve accuracies of $0.47$ and $0.42$ on this dataset~\citep{hendrycks2020many}. These methods use data augmentation techniques that are very relevant to the type of distribution shift between ImageNet and Imagenet-R. In the experiments in this paper, we only employ standard augmentations, i.e., random crop and random flip, and we make use of the unlabeled examples from this domain in the adaptation phase, hence our results are not comparable with the existing results reported on this dataset.

As shown in Table~\ref{tab:results_natural_shift} on datasets with natural shift, both iterative self-training and GIFT improve the accuracy over the best fine-tuned models. While in these scenarios there seems to be no big advantage for GIFT over iterative self-training, GIFT does not rely on the Assumption (a) that the target distribution should include samples from intermediate steps.

\subsection{Benchmarks with Synthetic Perturbations}
\begin{table*}[t]
    \centering
    \caption{Results of different adaptation techniques on perturbations of CIFAR10 in terms of accuracy on the test set, with WideResnet18-10. The total number of training steps is 1000 with a batch size of 512. The number of self-training iterations are 5 and 20 for iterative self-training, and \TheMethod respectively. None refers to the zero shot performance of the pretrained model. In all cases, \TheMethod outperforms all the baselines. 
    }
    \vspace{-10pt}
    \label{tab:cifar10:1000}
    \vskip 0.15in
    \begin{small}
    \begin{sc}
    \adjustbox{width=0.9\textwidth}{
    \begin{tabular}{l c c c c}
        \toprule
        \textbf{Target Domain} & \textbf{None}  & \textbf{self-training} &  \textbf{iterative self-training}  & \textbf{\TheMethod} \\\midrule
        Rotated CIFAR10 & 0.38 & 0.406 & 0.396 & \textbf{0.436}\\ 
        Scaled CIFAR10 & 0.559 & 0.558 & 0.578 & \textbf{0.615}\\
        Translated(0\%-100\%) CIFAR10 & 0.551 & 0.676 & 0.808 & \textbf{0.859} \\
        Translated(50\%-100\%) CIFAR10 & 0.262 & 0.421 & 0.658  & \textbf{0.729} \\
        Blurred CIFAR10 & 0.351 & 0.355 & 0.311 & \textbf{0.545} \\
        \bottomrule
    \end{tabular}}
    \end{sc}
    \end{small}
\end{table*}

To further investigate the effect of the type and degree of shift on the success of iterative self-training and GIFT we compare their performance on a synthetic benchmark where the target domain is created by applying synthetic perturbations on examples from the source domain, such as applying noise, rotating, scaling or translating images. 

To create these benchmarks, we split the training set of CIFAR10~\citep{krizhevsky2009learning} into equal-sized splits, where each split contains examples with different degrees of perturbation.
We use four types of perturbation:  rotation, scale, translate, and blur.
For rotation, we split the training data into three parts. 
In the first split images have a rotation angle of 0 to 5 degrees. In the second split images have a rotation angle of 5 to 55 degrees, and in the last split images have a rotation angle of 55 to 60. We use the first split $0\_5$ as the source domain and the third split $55\_60$ as the target domain.
For scale, translate, and blur we split the training data into two splits, where we use the first part with no perturbation as the source domain, and we apply the perturbation on the second part to get the out-of-distribution target domain. For both scale and translate we have no variation in the source and some variation in the target. For blur we only have one degree of blurring perturbation in the target and none in the source domain. 

\begin{figure*}[ht]
\centering
    \captionsetup[subfigure]{oneside,margin={0cm,0cm}}
    \subfloat[Accuracy]{
        \includegraphics[width=0.999\textwidth]{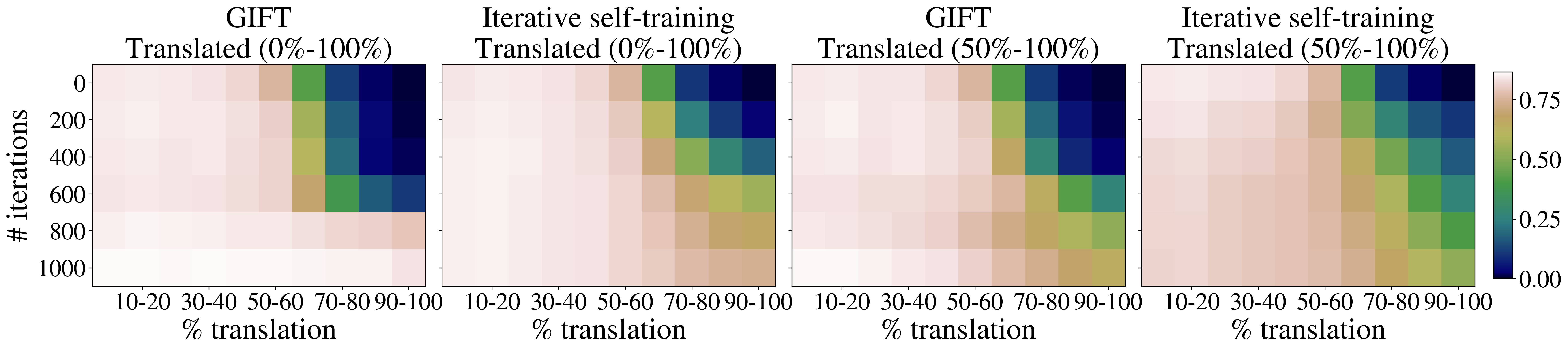} 
}\\\vspace{-5pt}

 \subfloat[Confidence]{
    \includegraphics[width=0.999\textwidth]{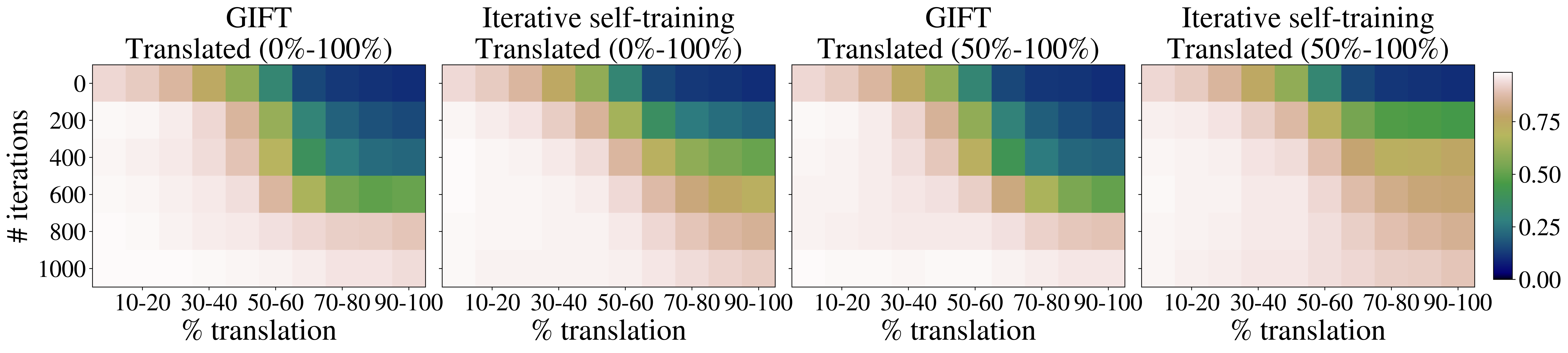}
}
    \caption{Accuracy and Confidence of \TheMethod and iterative self-training as a function of the number of training iterations on different bins of the translated CIFAR10. The accuracy is evaluated for the test data with translations between 0 and 100\% (where zero means no shift, and 100\% means the maximum possible amount of shift, i.e., $\textsc{image width}/2$). The iterative self-training model has 5 teacher updates over 1000 training iterations (i.e. updates happen after 0, 200, 400, 600, 800 iterations). GIFT has 20 teacher updates over 1000 iterations (i.e. updates happen after 0, 50, 100, ..., 950 iterations). The left two panels in each row correspond to models that are trained for the target dataset Translated (0\%-100\%) CIFAR10. The right two panels in each row correspond to models trained for the target dataset Translated (50\%-100\%) CIFAR10.}
    \label{fig:curriculum_ablation}
    \vspace{-10pt}
\end{figure*}

The results on CIFAR10 with different perturbations are shown in Tables~\ref{tab:cifar10:1000} and \ref{tab:cifar10_500}. They indicate the superiority of \TheMethod to iterative self-training. The advantage of GIFT over iterative-self-training is more apparent when the target distribution does not include a diverse set of samples (Translated vs Blur), or when the two distributions are not overlapping (Translated(0\%-100\%) vs Translated(50\%-100\%)).
Additionally, compared to GIFT the performance of iterative self-training drops more significantly when decreasing the total number of training steps. We can see this by contrasting the results in Table~\ref{tab:cifar10:1000} and \ref{tab:cifar10_500}.

\begin{table}[ht]
    \centering
    \caption{Results of different adaptation techniques for WideResnet18-10 trained on perturbations of CIFAR10 in terms of accuracy on the out-of-distribution set. The total number of training steps is 500 with a batch size of 512. The number of self-training iterations are 2 and 20 for iterative self-training and GIFT respectively. Comparing this results to Table~\ref{tab:cifar10:1000} where the total number of training steps is 1000, we see a noticeable drop in the performance of iterative self-training while the results for GIFT are more stable.} 
    \label{tab:cifar10_500}
    \begin{small}
    \begin{sc}
    \adjustbox{width=0.6\textwidth}{
    \begin{tabular}{l c c}
        \toprule
        \textbf{Target Domain} & \textbf{iterative self-training} & \textbf{GIFT} \\ \midrule
         Rotated Cifar10 & 0.408 & 0.4176 \\
         Scaled CIFAR10 & 0.573  &  0.6767\\
         Translated(0\%-100\%) CIFAR10 & 0.629 & 0.8349\\
         Translated(50\%-100\%) CIFAR10 & 0.477 & 0.8319 \\
         Blurred CIFAR10 & 0.3551  & 0.5213 \\
         \bottomrule
    \end{tabular}}
    \end{sc}
    \end{small}
\end{table}

\begin{figure*}[t!]%
    \centering
    \captionsetup[subfigure]{oneside,margin={-5cm,0cm}}
    \subfloat[Iterative self-training]{{
    \begin{tikzpicture}
    \begin{axis}[
        width=0.4\textwidth, height=4cm,
        xlabel=Number of teacher updates,
        ylabel=Accuracy,
        grid style=dashed,
        ymin=0.08,ymax=0.9,
        xtick=data,
        ymajorgrids,
        xmajorgrids,
        xmode=log,
        xtick={1, 2, 5, 10, 20},
        xticklabels={1, 2, 5, 10, 20},
         legend style={
            at={(1.4,1.2)},
            anchor=north,
            legend columns=6,
            inner sep=0.2pt,
            outer sep=0.2pt,
            font=\fontsize{7}{4}\selectfont,
    },
    legend cell align={left},
    ]
    \addplot table[x index=0,y index=1,] {cifar10_scale_1000_direct.dat};
    \addlegendentry{Scaled}
    
    \addplot table[x index=0,y index=1,] {cifar10_translate_1000_direct.dat};
    \addlegendentry{Translated(0\%-100\%)}
    
    \addplot table[x index=0,y index=1,] {cifar10_blur_1000_direct.dat};
    \addlegendentry{Blurred}
    
    \addplot table[x index=0,y index=1,] {cifar10_translate_gap_1000_direct.dat};
    \addlegendentry{Translated(50\%-100\%)}
    
    \addplot table[x index=0,y index=1,] {cifar10_mixed_1000_direct.dat};
    \addlegendentry{Mixed}
    
    \end{axis}
\end{tikzpicture}%
    } %
    \label{fig:cifar10_1000_direct}}   \qquad
    \hspace{-180pt}
    \captionsetup[subfigure]{oneside,margin={0cm,0cm}}
\subfloat[GIFT]
    {{
    \begin{tikzpicture}
    \begin{axis}[
        width=0.4\textwidth, height=4cm,
        xlabel=Number of teacher updates,
        ylabel=Accuracy,
        grid style=dashed,
        ymin=0.08,ymax=0.9,
        xtick=data,
        ymajorgrids,
        xmajorgrids,
        xmode=log,
        xtick={5, 10, 20, 40},
        xticklabels={5, 10, 20, 40},
    ]
    \addplot table[x index=0,y index=1,] {cifar10_scale_1000_gift.dat};
    
    \addplot table[x index=0,y index=1,] {cifar10_translate_1000_gift.dat};
    
    \addplot table[x index=0,y index=1,] {cifar10_blur_1000_gift.dat};
    
    \addplot table[x index=0,y index=1,] {cifar10_translate_gap_1000_gift.dat};
    
    \addplot table[x index=0,y index=1,] {cifar10_mixed_1000_gift.dat};
    
    \end{axis}
    
\end{tikzpicture}%
    } %
        \label{fig:cifar10_1000_gift}
    }

    \qquad
    \caption{Accuracy of iterative self-training and \TheMethod on perturbations of CIFAR10 as a function of the number of teacher updates when total number of training steps is 1000. Accuracies of both models improves by increasing the number of self training iterations up to a threshold. Beyond the threshold, while iterative self-training performance deteriorates, \TheMethod saturates and hence shows more robustness.}
    \label{fig:cifar10:iterations_1000}
\end{figure*}
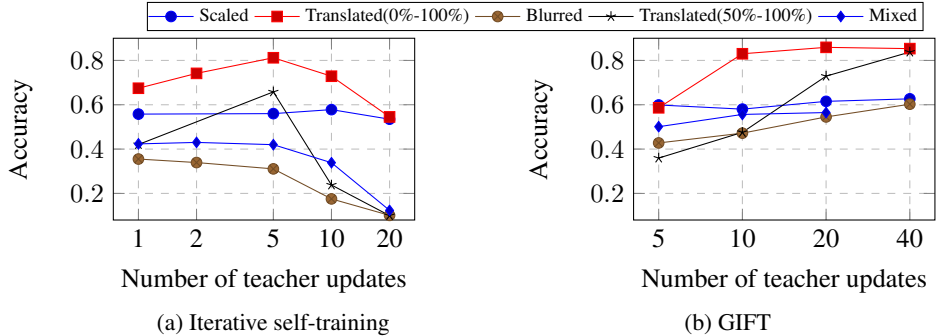

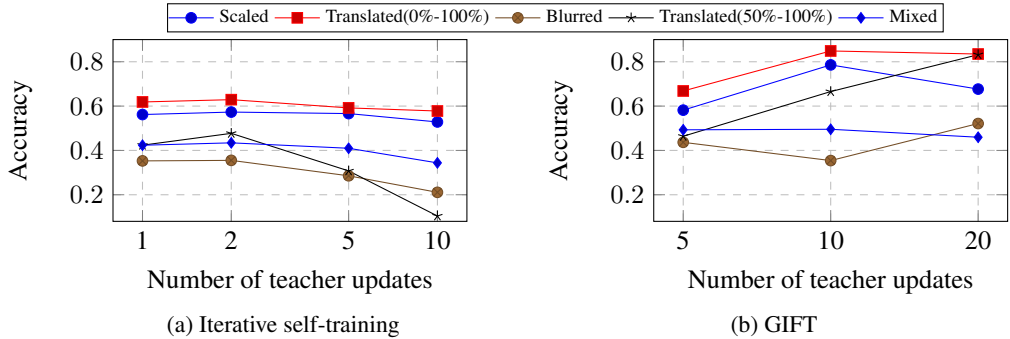
\begin{figure*}[ht]%
    \centering
    \captionsetup[subfigure]{oneside,margin={-5cm,0cm}}
    \subfloat[Iterative self-training]{{
    \begin{tikzpicture}
    \begin{axis}[
        width=0.45\textwidth, height=4cm,
        xlabel=Number of teacher updates,
        ylabel=Accuracy,
        grid style=dashed,
        ymin=0.08,ymax=0.9,
        xtick=data,
        ymajorgrids,
        xmajorgrids,
        xmode=log,
        xtick={1, 2, 5, 10, 20},
        xticklabels={1, 2, 5, 10, 20},
        legend style={
            at={(1.25,1.2)},
            anchor=north,
            legend columns=6,
            inner sep=0.2pt,
            outer sep=0.2pt,
            font=\fontsize{7}{4}\selectfont,
        },
        legend cell align={left},
    ]
    
    \addplot table[x index=0,y index=1,] {cifar10_scale_500_direct.dat};
    \addlegendentry{Scaled}
    
    \addplot table[x index=0,y index=1,] {cifar10_translate_500_direct.dat};
    \addlegendentry{Translated(0\%-100\%)}
    
    \addplot table[x index=0,y index=1,] {cifar10_blur_500_direct.dat};
    \addlegendentry{Blurred}
    
    \addplot table[x index=0,y index=1,] {cifar10_translate_gap_500_direct.dat};
    \addlegendentry{Translated(50\%-100\%)}
    
    \addplot table[x index=0,y index=1,] {cifar10_mixed_500_direct.dat};
    \addlegendentry{Mixed}
    
    \end{axis}
\end{tikzpicture}%
    } %
    \label{fig:cifar10_500_direct}}   \qquad
    \hspace{-180pt}
    \captionsetup[subfigure]{oneside,margin={0cm,0cm}}
\subfloat[GIFT]
    {{
    \begin{tikzpicture}
    \begin{axis}[
        width=0.45\textwidth, height=4cm,
        xlabel=Number of teacher updates,
        ylabel=Accuracy,
        grid style=dashed,
        ymin=0.08,ymax=0.9,
        xtick=data,
        ymajorgrids,
        xmajorgrids,
        xmode=log,
        xtick={5, 10, 20, 40},
        xticklabels={5, 10, 20, 40},
    ]
    \addplot table[x index=0,y index=1,] {cifar10_scale_500_gift.dat};
    
    \addplot table[x index=0,y index=1,] {cifar10_translate_500_gift.dat};
    
    \addplot table[x index=0,y index=1,] {cifar10_blur_500_gift.dat};
    
    \addplot table[x index=0,y index=1,] {cifar10_translate_gap_500_gift.dat};
    
    \addplot table[x index=0,y index=1,] {cifar10_mixed_500_gift.dat};
    
    \end{axis}
\end{tikzpicture}%
    } %
    \label{fig:cifar10_500_gift}
    }
    \qquad
    \caption{Accuracy of iterative self-training and \TheMethod on perturbations of CIFAR10  as a function of the number of teacher updates when total number of training steps is 500. We see that the benefit of using GIFT over iterative self-training is more when we have fewer number of training steps. Not only there is a smaller drop in the accuracy of the models trained with \TheMethod, but also it is more robust with respect to variations in the number of teacher updates.
    \label{fig:cifar10:iterations_500}
    }
    \vspace{-15pt}
\end{figure*}

\paragraph{Investigating the Curricula:}
To empirically confirm the hypothesis that both iterative self-training and \TheMethod gradually guide the model to fit the out-of-distribution target distribution,  we track the accuracy and confidence of the models on different subsets of the target data throughout the training. This is shown in in Figures~\ref{fig:curriculum_ablation}. We see that for both methods the accuracy and confidence measures are increasing incrementally from easier examples, i.e., examples with smaller amounts of perturbation, to harder examples, i.e., examples with larger amounts of perturbation. 

For \TheMethod the number of iterations is tied to the steps in which we increment  the interpolation coefficient, $\lambda$, and it is interesting to see that there is a correlation between $\lambda$ and the accuracy and confidence of the model on actual intermediate steps, as shown in Figure~\ref{fig:curriculum_ablation}. 
For iterative self-training, during training the model selects training examples for which to compute the loss based on its confidence. 
Hence, tracking the confidence as a function of translation percentage during training allows us to see whether the model is indeed selecting less perturbed examples earlier and more perturbed examples later in the adaptation phase and thus creates its own curriculum.
As we see in Figure~\ref{fig:curriculum_ablation}, for both iterative self-training and \TheMethod the confidence of the models decreases gradually for data with an increased level of perturbation. Note also that the confidence scores show very similar patterns to the accuracies.
This confirms that the reason behind the success of iterative self-training is the implicit curriculum strategy, and that the gradual interpolation strategy that we employ in \TheMethod can be a good proxy for gradual self-training when we do not have access to the gradually shifted data.

\paragraph{Effect of number of intermediate steps:}
We examine the effect of the number of intermediate steps when the total number of training steps is fixed. As illustrated in Figures~\ref{fig:cifar10:iterations_1000} and \ref{fig:cifar10:iterations_500}, increasing the number of intermediate steps (teacher updates), which means taking smaller steps in the gradual adaptation procedure, leads to a better performance up to a threshold for \TheMethod and iterative self-training. If we keep increasing the number of intermediate steps, the performance of the models  decreases rapidly for iterative self-training, whereas \TheMethod is more robust with respect to the number of intermediate steps.
This decrease in the performance is potentially due to accumulations of the errors of the self-training process or because with a fixed number of training steps, increasing the number of intermediate steps leads to a decrease in the number of iterations in each self-training step, which could mean the model can not adapt well to each intermediate step. The robustness of the \TheMethod, in this case, could mean that this method needs fewer number of iterations in each intermediate step.
Comparing Figures~\ref{fig:cifar10:iterations_1000} and ~\ref{fig:cifar10:iterations_500}, we see that for both iterative self-training and GIFT, decreasing the number of total training steps (500 vs 1000), reduces the effective number of teacher updates. This confirms the hypothesis that the model can benefit from more teacher updates if it has enough time to properly adapt to each intermediate step. We provide more analysis for this in Appendix~\ref{app:steps}.

\section{Related Work}
\paragraph{Unsupervised domain adaptation:}
In unsupervised domain adaptation we have access to labeled examples from the source domain(s) and unlabeled examples from the target domain(s) and the goal is to get a good performance on the target domain.
Unsupervised domain adaptation techniques fall within three main categories~\citep{sun2020unsupervised}:
(1) Methods based on matching the feature distributions of source and target domains. Algorithms in this group rely on the assumption that the models can learn domain invariant representations, and they employ different self-supervised based losses to enforce this invariance by exposing the model to the unlabeled data from the target distribution~\citep{ganin2016domain, ben:rep_analysis}. 
(2) Methods based on transforming source and target distributions. They analyze the input space and project source and target data to a lower dimensional manifold and try to find a transformation between the two~\citep{fernando2013unsupervised, gopalan2011domain, harel2010learning}. Another approach in this line of work is to transform the source data to be as close to the target data as possible. For example, \citet{sun2017correlation} matches the second order statistics of the input spaces.
(3) Self-training based methods. We discuss these in the next paragraph.

\paragraph{Self-training:}
Recent works have shown significant progress using self-training in computer vision \citep{xie2019self, yalniz2019billion, zoph2020rethinking}. Self-training has also been used for domain adaptation by generating pseudo labels in the target domain and directly training a model for the target domain \citep{xie2018learning, saito2017asymmetric, chang2019domain, manders2018simple, zou2019confidence, zou2018unsupervised}. \citet{xie2018learning} align labeled source centroids and pseudo-labeled target centroids. \citep{chang2019domain} uses different normalization parameters for source examples and pseudo labeled examples in the target domain. \citet{zou2019confidence} introduced label-regularized self-training which generates soft pseudo-labels for self-training. Different from these works, we use a curriculum learning approach where we generate pseudo labels for intermediate virtual examples and gradually adapt the model to the target domain.
\paragraph{Curriculum learning:} 
Curriculum learning~\citep{elman1993learning,sanger1994neural, bengio2009curriculum} has led to better performance in terms of generalization and/or convergence speed in many domains such as computer
vision \citep{pentina2015curriculum,sarafianos2017curriculum,Guo_2018_ECCV, Wang_2019_ICCV},  natural language processing \citep{cirik2016visualizing,platanios2019competence}  and neural
evolutionary computing~\citep{zaremba2014learning}. On the other hand, there has been some negative results in neural machine translation~\citep{kocmi2017curriculum, zhang2018empirical, zhang2019curriculum}.
In this work, we investigate it in the unsupervised domain adaptation scenario. Different notions of ``difficulty'' of examples are used in the literature, such as using the loss value of a pre-trained model~\citep{bengio2009curriculum}, or
the first iteration in which an example is learned and remains learned after that~\citep{Forgetting}. \citet{jiang2020exploring} have proposed using a consistency score calculated based on the consistency of a model in correctly predicting a particular example’s label trained on i.i.d. draws from the training set. \citet{wu2020curricula} show that all these difficulty scores are consistent. Here, we use the coefficient of linear interpolation between source and target representation as a measure of difficulty of a (virtual) sample.


\section{Discussion and Conclusion}
We show the importance of having a proper curriculum for the success of self-training algorithms for domain adaptation. We first demonstrate that applying self-training iteratively can successfully adapt the model to the new target distribution if the target distribution contains the intermediate examples, and that this method is less effective if the source and target distributions are not overlapping. GIFT, the method that we propose, is specifically designed to deal with cases where there is a gap between distributions, where plainly applying self-training iteratively and filtering examples based on the confidence of the model would not result in a proper curriculum, simply because the intermediate examples are missing. 

We report results on a set of natural distribution shift benchmarks.
Additionally, we have designed experiments on a synthetic benchmark created from CIFAR10. This allows us to control the properties of the domain shift, so that we can improve our understanding of how different self-training based approaches are affected by different domain shift settings. More specifically, we have designed our experiments such that we can control the amount of shift, the type of shift, and whether there is a gap between the data distributions in the source and target domain. 

GIFT is inspired by existing works that rely on interpolating representations in the input or feature space~\citep{gong2012geodesic,verma2019manifold,ijcai2019-504}, however, we do not provide  theoretical guarantees that our approach for generating virtual samples based on simple convex interpolations and Sinkhorn or pseudo random alignment would lead to better adaptation. An interesting direction for future work could be to investigate other ways of interpolating in the feature space and explore the use of more sophisticated interpolation schemes,  for instance schemes based on optimal transport. Moreover, the experiments in this paper are limited to image classification. Extending our approach beyond classification and to other data modalities such as textual data is an important next step. 

\subsection*{Acknowledgement}
We would like to thank Behnam Neyshabur for useful discussions and his feedback on the paper.
We appreciate conversations with Lasse Espeholt, Hossein Mobahi, Avital Oliver and Jeremiah Liu along the way.

\bibliography{inertia}
\bibliographystyle{plainnat}

\newpage

\clearpage
\appendix

\title{Supplementary Material for ``Gradual Domain Adaptation in the Absence of Intermediate Distributions''}
\date{}
\author{}
\maketitle


\section{Experimental Setup}
\label{app:exp_setup}

\paragraph{Optimizer and learning rate schedule.} In all our experiments on datasets with natural distribution shift, we use Adam optimizer. When training on the labeled source domain we use a learning rate schedule with cosine  decay and initial learning rate of 1e-4 and when adapting to the target domain we use a learning rate of 1e-5 with exponential decay rate of 0.9. 
For pretraining the models on ImageNet-1k, we use a batch size of 1024 and the learning rate schedule is linear warmup (for 5 epochs) + cosine decay with the base learning rate of 0.1.

For Perturbed Cifar10 experiments, we use SGD with momemtum. During the pretraining stage on the source domain, the learning rate schedule is cosine decay with an initial learning rate of 0.1. In all experiments we use L2 loss with the weight of 1e-5. During the adaptation phase we use a batch size of 256 and the learning rate is constant and set to 1e-3. 
For experiments on Perturbed Cifar10 with higher number of training steps (20000 steps), we use a lower learning rate of 1e-4 in the adaptation phase. 

\paragraph{Neural network architectures.} 
 For experiments on dataset with natural shift, we use a ResNet101.
For the experiments on Perturbed Cifar10, we use a WideResNet28-10 with a dropout rate of 0.3.
For the virtual interpolations in \TheMethod: we use the first three layers (input, initial convolution layer, and the layer above it).

\paragraph{Training on source during the adaptation phase.} We do not train the model on the source data during the adaptation phase. While in some cases this could result in a better performance on both the source and target domain, our assumption here is that there is no reason for the source and target data to be compatible, i.e. it is possible for the model to not be able to fit both distributions simultaneously.

\paragraph{Regularization}
During adaptation, we use the weight decay of 0.01.
In addition to weight decay, we use another regularization term that encourages the model to stay close to its initial state. This regularization term is simply computed as the L2 distance of the parameters of the model and their value at its initial state. We set the weight for this factor to 0.001 in all adaptation experiments.

\paragraph{Backpropagating gradients through interpolations.} We train the model with the representations of virtual examples that we create by interpolating the representations from real examples. During pretraining on the source domain we use manifold mixup regularization, where we interpolate between representations of labeled source examples. During the adaptation stage that is part of \TheMethod we interpolate between labeled source representations and unlabeled target representations. One important hyper-parameter related to this is whether in the backward pass we back-propagate the gradients all the way down to the input layer or stop the gradients at the layer in which the interpolated representations are computed. In our experiments, similar to \citet{verma2019manifold} we allow the gradients to pass through the interpolated activations.


\paragraph{Computational Resources:} We train our models on cloud TPU devices. Our estimate of the amount of compute used for the experiments of this paper is roughly about 2k TPU-core days. 

\section{Training on Labeled Source Domain}
\label{app:reg}

We compare four different approaches for training the model on labeled source data.

\paragraph{Standard fine-tuning:} Given labeled examples from a source domain, we employ a model that is pretrained on some large scale dataset, Imagenet-1k  in our  case, replace its head (projection layer) with a new head for the task at hand, and update all its parameters to fit the source domain  data.
\paragraph{Mixup with convex combination interpolations:} During fine-tuning on labeled source data, we apply mixup/manifold~\citep{zhang2018mixup, verma2019manifold}  on the input and activation from the first layer of  model, and  to compute the interpolations we simply compute  the convex combination of features for two randomly aligned  examples.
\paragraph{Mixup with wasserstein interpolations:} During fine-tuning on labeled source data, we apply a variant of mixup/manifold mixup where interpolations are computed using the closed form Monge Map for Gaussian Wasserstein distances. In our experiments we observed that in some cases, interpolating examples in this alternative way leads to better results compared to the convex interpolations used in \citet{verma2019manifold}. 
\paragraph{Domain advarserial neural networks (DANN):}
Given labeled samples from a source domain and unlabeled samples from target domain, DANN~\citep{ganin2016domain} learns domain invariant representations, while minimizing its prediction error on labels source data.
In our experiments the output of  the prelogits  layer is fed to the  domain classifier, and  the scheduling of the weight of the domain classification loss  is the same as what is suggested in \citet{ganin2016domain}.


\subsection{Manifold Mixup with Wasserstein Interpolation}
During training on the labeled source dataset we use a variant of manifold mixup~\citep{verma2019manifold} with an adapted strategy for interpolation.
In manifold mixup~\citep{verma2019manifold}, representations of virtual examples are created by linearly interpolating representations of two randomly aligned examples $(x_i, x_j)$ in a randomly selected layer $L$ of the neural network $M_\theta$. The labels for the interpolated examples, $\hat{y}_{ij}$, are computed by interpolating the labels of the aligned examples, $(y_i, y_j)$, using the same interpolation coefficient, $\lambda$. This is summarized in \eqref{eq:manifoldmizup}.
\begin{equation} \label{eq:manifoldmizup}
\begin{split}
    \hat{z}_{ij} = & (1 - \lambda) \mathcal{M}^{:L}_\theta(x_i) + \lambda \mathcal{M}^{:L}_\theta(x_j) \\
    \hat{y}_{ij} = & (1 - \lambda) y_i + \lambda y_j \\
    \end{split}.
\end{equation}
Here $M_\theta^L$ denotes the part of the neural network that outputs the activations of layer $L$.
The interpolated representations are then fed into the rest of the neural network at layer $L$, and together with the interpolated labels they serve as additional `data' to which the model is fit.


In our experiments we take a different approach to interpolation. Inspired by the style transfer method discussed in \citet{pmlr-v108-mroueh20a}, we use interpolations based on the Wasserstein distance between two Gaussian distributions that are fit to representations of two input images. i.e., the spatial features in the representations of two images are the datapoints for two datasets. We estimate the empirical mean and diagonal covariance matrices for these datapoints and use the closed form optimal transport map between two Gaussian distributions to interpolate the spatial features between two representations. More precisely, given two images $x_i$ and $x_j$, we compute representations $z_i = M_{\theta}^L(x_i)$ and $z_j = M_{\theta}^L(x_j)$, where $z_i$ and $z_j$ are three-dimensional tensors with a width $W^L$, height $H^L$ and channel size $C^L$. Each spatial feature vector of size $C^L$ within $z_i$ and $z_j$ is considered one datapoint. We compute the average feature vectors within $z_i$ and $z_j$, denoted by $\mu_i$ and $ \mu_j$ respectively, as well as the variances $ \sigma_i^2$ and $ \sigma_j^2$. Note that we are approximating the empirical covariance matrices with diagonal matrices with the variances $\sigma_i^2$ and $ \sigma_j^2$ on the diagonals. Given these quantities, we can compute the closed form Monge map between two Gaussian distributions with diagonal covariance matrices as 
\begin{align}
    T_{z_i\rightarrow z_j}(z) = \mu_j + \mathrm{diag}\left(\frac{ \sigma_j}{\sigma_i}\right) ( z - \mu_i).
\end{align}
Here $z$ is understood to be a feature map of the same size as $z_i$ and $z_j$. Interpolated representations are then computed with
\begin{align}
    \hat{z}_{ij} = (1-\lambda) z_i + \lambda T_{z_i\rightarrow z_j}(z_i).
\end{align}
Similar to its use in style transfer \citep{pmlr-v108-mroueh20a}, we assume this transformation does not change the content of the representation, and we therefore do not apply an interpolation scheme to the labels $y_i$ and $y_j$ of datapoints $x_i$ and $x_j$. Instead, we use the label $\hat{y}_{ij} = y_i$ for the virtual representation $\hat z_{ij}$.




In \citet{verma2019manifold} the interpolation coefficient $\lambda$ is sampled from a Beta distribution $\mathrm{Beta}(\alpha,\beta)$, where $\alpha$ and $\beta$ are hyperparameters. In our experiments we set both $\alpha$ and $\beta$ to 1, so that we are effectively sampling $\lambda$ uniformly from the interval $[0,1)$.

\section{Effect of Number of Teacher Updates}
\label{app:steps}
\begin{figure}[h]%
    \centering
    \captionsetup[subfigure]{oneside,margin={-3cm,0cm}}
    \subfloat[iterative self-training]{{
    \begin{tikzpicture}
    \begin{axis}[
         width=0.4\textwidth, height=5cm,
        xlabel=Number of teacher updates,
        ylabel=Accuracy,
        grid style=dashed,
        ymin=0.08,ymax=0.9,
        xtick=data,
        ymajorgrids,
        xmajorgrids,
        xmode=log,
        xtick={1, 2, 5, 10, 20, 40},
        xticklabels={1, 2, 5, 10, 20, 40},
        legend style={
            at={(1.25,1.15)},
            anchor=north,
            legend columns=6,
            inner sep=0.2pt,
            outer sep=0.2pt,
            font=\fontsize{7}{4}\selectfont,
    },
    legend cell align={left},
    ]
    
    \addplot table[x index=0,y index=1,] {cifar10_translate_direct_step_100.dat};
    \addlegendentry{Translated (0\%-100\%)}
    
    
    \addplot table[x index=0,y index=1,] {cifar10_translate_gap_direct_step_100.dat};
    \addlegendentry{Translated(50\%-100\%)}
    
     \addplot table[x index=0,y index=1,] {cifar10_blur_direct_step_100.dat};
    \addlegendentry{Blurred}
    
    \end{axis}
\end{tikzpicture}%
    } %
    \label{fig:cifar10_step_100_direct}}   \qquad
    \hspace{-140pt}
    \captionsetup[subfigure]{oneside,margin={0cm,0cm}}
    \subfloat[GIFT]{{
    \begin{tikzpicture}
    \begin{axis}[
         width=0.4\textwidth, height=5cm,
        xlabel=Number of teacher updates,
        ylabel=Accuracy,
        grid style=dashed,
        ymin=0.08,ymax=0.9,
        xtick=data,
        ymajorgrids,
        xmajorgrids,
        xmode=log,
        xtick={5, 10, 20, 40},
        xticklabels={5, 10, 20, 40},
    ]
    
    \addplot table[x index=0,y index=1,] {cifar10_translate_gift_step_100.dat};
    
    
    \addplot table[x index=0,y index=1,] {cifar10_translate_gap_gift_step_100.dat};
    
    \addplot table[x index=0,y index=1,] {cifar10_blur_gift_step_100.dat};
    
    \end{axis}
\end{tikzpicture}%
    } %
    \label{fig:cifar10_step100_gift}}  
\caption{Effect of the number of teacher updates on the accuracy when the number of training steps before each teacher update is fixed and set to 100, for different perturbations of CIFAR10. For Translated (0\%-100\%) CIFAR10 and Translated (50\%-100\%) CIFAR10, we see an increasing trend in accuracy as we increase the number of teacher updates for both iterative self-training and \TheMethod. Whereas for Blurred CIFAR10, the accuracy decreases for iterative self-training. 
} 
\label{fig:cifar10:iterations_step100}
\end{figure}
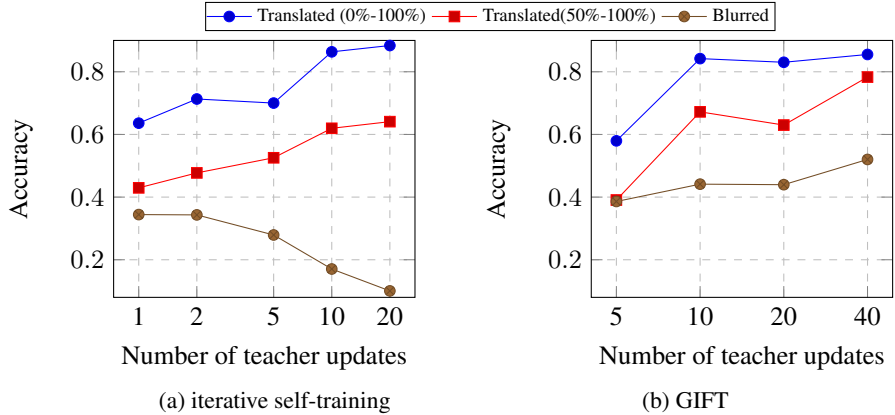

\begin{figure}[ht]%
    \centering
    \begin{tikzpicture}
    \begin{axis}[
        width=0.4\textwidth, height=5cm,
        xlabel=Number of teacher updates,
        ylabel=Accuracy,
        grid style=dashed,
        ymin=0.08,ymax=0.9,
        xtick=data,
        ymajorgrids,
        xmajorgrids,
        xmode=log,
        xtick={1, 5, 10, 20, 40},
        xticklabels={1, 5, 10, 20, 40},
         legend style={
            at={(1.5,.8)},
            anchor=north,
            legend columns=1,
            inner sep=0.2pt,
            outer sep=0.2pt,
            font=\fontsize{7}{4}\selectfont,
    },
    legend cell align={left},
    ]
    \addplot table[x index=0,y index=1,] {cifar10_scale_1000_interpolated.dat};
    \addlegendentry{Scaled}
    
    \addplot table[x index=0,y index=1,] {cifar10_translate_1000_interpolated.dat};
    \addlegendentry{Translated (0\%-100\%)}
    
    \addplot table[x index=0,y index=1,] {cifar10_blur_1000_interpolated.dat};
    \addlegendentry{Blurred}
    
    \addplot table[x index=0,y index=1,] {cifar10_translate_gap_1000_interpolated.dat};
    \addlegendentry{Translated(50\%-100\%)}
    
    \end{axis}
\end{tikzpicture}%
\caption{Effect of number of self training iterations on accuracy when all interpolations are represented to the model at the same time with the total number of training steps of 1000 for different perturbations of CIFAR10. Similar to iterative self-training, the performance improves by increasing the number of self training iterations up to a threshold. Beyond the threshold, the performance deteriorates.} 
\label{fig:cifar10:iterations_1000_interpolated}
\end{figure}
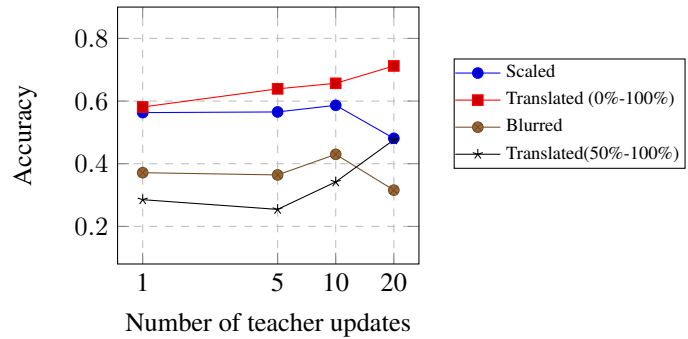

To better distinguish the effect of the number of teacher updates from the number of training steps between each two consecutive teacher updates, in Figure~\ref{fig:cifar10:iterations_step100} we plot the accuracy as a function of the number of teacher updates when the number of training steps is fixed, i.e., the total number of training steps increases as we increase the number of teacher updates. For \TheMethod, we observe an increasing trend in the accuracy as the number of teacher update increases on all the benchmarks. However, for iterative self-training we only see a benefit in increasing the number of teacher updates for datasets with a range of perturbations in the target domain such as the Translated CIFAR10 datasets, as opposed to Blurred CIFAR10.

Additionally, Figure~\ref{fig:cifar10:iterations_1000_interpolated} shows the effect of the number of teacher updates when the total number training steps is 1000 for a non gradual version of \TheMethod, where all the interpolations are presented to the model simultaneously. We observe that compared to \TheMethod, where the value of the interpolation coefficient $\lambda$ is gradually increased, having more teacher updates is much less beneficial.

\end{document}